# Explorative analysis of human disease-symptoms relations using the Convolutional Neural Network

**Zolzaya Dashdorj [1,2]*, Stanislav Grigorev [2] and Munguntsatsral Dovdondash [1]**

[1] Mongolian University of Science and Technology; zolzaya@must.edu.mn, munguntsatsral.do@gmail.com
[2] Irkutsk National Research Technical University; svg@istu.edu
* Correspondence: zolzaya@must.edu.mn;

## ABSTRACT

In the field of health-care and bio-medical research, understanding the relationship between the symptoms of diseases is crucial for early diagnosis and determining hidden relationships between diseases. The study aimed to understand the extent of symptom types in disease prediction tasks. In this research, we analyze a pre-generated symptom-based human disease dataset and demonstrate the degree of predictability for each disease based on the Convolutional Neural Network and the Support Vector Machine. Ambiguity of disease is studied using the K-Means and the Principal Component Analysis. Our results indicate that machine learning can potentially diagnose diseases with the 98-100% accuracy in the early stage, taking the characteristics of symptoms into account. Our result highlights that types of unusual symptoms are a good proxy for disease early identification accurately. We also highlight that unusual symptoms increase the accuracy of the disease prediction task.

## INTRODUCTION

The past decades have brought remarkable advances in our understanding of human disease [1,10]. In Mongolia, the rural population for 2021 was 1,045,010, a 1.21% increase from 2020, that is 31% of the total population. Due to insufficient doctors, nurses in rural areas, providing primary healthcare is a real challenge. The ability of artificial intelligence to process thousands of pages of clinical notes per second in search of the necessary information could provide the essential data that allows us to achieve outstanding results in diagnosing various types of diseases. Many applications have been developed, including health-care chatbots, and disease diagnosis by CT and MRI images. Symptoms are essential predictors to diagnose diseases and are commonly used in the early stage or during treatment [1-9]. A recent trend in health-care diagnosis research employs machine learning techniques [2-8]. [2] a weighted KNN algorithm was used to identify a disease based on the symptoms, age, and gender of an individual. The accuracy of the weighted KNN algorithm for the prediction was 93.5 %, approximately on 230 disease types. A particular disease, such as Parkinson's disease has been studied based on motor and non-motor symptoms and other symptoms like memory disorders, olfactory disorder, sleep disorders and many more [4]. Those symptoms were collected in the form of signals, images, videos, or clinical measures from the articles published between 2017 and 2019. Most of the work has been done on a smaller dataset. Large datasets are needed for generalization. Network analysis is helpful in many research and applications in terms of large-scale visualizations. [1] found that the symptom-based similarity of two diseases correlates strongly with the number of shared genetic associations and the extent to which their associated proteins interact. More specifically, [5] studied the predictability of heart disease using the Multilayer Perceptron Neural Network. Essential 76 characteristics describing heart health, such as age, gender and pulse rate, are collected from the UCI Cleveland Library dataset. The average prediction was 91% of precision and 89% of recall. Similarly, diabetes [6] is a growing chronic, life-threatening disease that affects millions of people. Five hundred twenty instances are collected using direct questionnaires from the patients. The prediction rate was estimated between 87.5% and 97.4% using Naive Bayes Algorithm, Logistic Regression Algorithm, and Random Forest Algorithm. [7] analyzes older adults to predict hospitalization and mortality. Self-reported symptoms were collected; common symptoms were musculoskeletal pain, fatigue, back pain, shortness of breath and difficulty sleeping. A summary score was observed as a predictor score of hospitalization and mortality. However, there are few studies to discover to what extent type of symptoms can be used to diagnose a particular disease. [11] studied clinical notes for mining disease and symptoms relations based on word embedding learned through neural networks. Related diseases and symptoms were observed by the suggested approach applied in 154,738 clinical notes. [12] proposed a comprehensive framework to impute missing symptom values by managing uncertainty present in the data set.In this study, we analyze a disease-symptom relation network to understand the characteristics of patients' symptoms in terms of occurrence to improve a disease diagnosis and prediction task. We demonstrate the study by estimating the disease predictability and the relation between diseases and symptoms using machine (deep) learning techniques (SVM and CNN) based on word embedding. Understanding the associations between the diseases and symptoms in clinical notes can support physicians in making decisions, and provide researchers evidence about disease development and treatment as well as primary care applications.

**METHODS AND MATERIALS**

The following 3-stage research was conducted using machine learning to understand the predictability of diseases given symptoms. We use the same dataset used in [8]. Those research studies obtained a 95.12% accuracy score in disease prediction tasks by employing Decision Tree, Random Forest and Naive Bayes classifiers. A total of 4,920 patient records were obtained in this research. The dataset of 41 types of disease consisting of 135 symptoms was used, and the degree of symptom severity was graded on three levels. Every disease in our dataset is associated with up to 18 symptoms.

1. Analyze common and unusual symptoms.

It is crucial to explore hidden links between diseases to understand their characteristic difference. Symptoms are commonly observed in diseases and are essential to diagnose disease primarily. We analyze common and unusual symptoms in the disease-symptom network and estimate the rate of the uniqueness of the symptoms by their occurrence.

2. Estimate the degree of predictability of disease based on symptoms.

Based on common and unusual symptoms, we attempt to diagnose disease in an early state based on Support Vector Machine and CNN algorithms. In the data preprocessing stage, we applied a bag of words of natural language processing methods, representing symptoms and disease types in free text.

Support Vector Machine (SVM). To calculate the inner product in a feature space of symptoms as a function of the main entry points, a nonlinear learning machine is built as a kernel function and can be expressed as K. A kernel function may be interpreted as a k function, and consequently, for all $x,z \in X$, we have:

$$K(x,z)=\langle\varphi(x)\cdot\varphi(z)\rangle \tag{1}$$

We use a radial basis function (RBF) as the least square SVM (LS-SVM) for SVM. The main advantage of LS-SVM is that it is more efficient than SVM in terms of computation, whereby LS-SVM training only solves a set of linear equations instead of the time-consuming and challenging calculation of second-order equations.

Convolutional neural network (CNN). In a neural network, neurons are fed inputs, which then neurons consider the weighted sum over them and pass it by an activation function and pass out the output to the next neuron. A CNN differs from that of a neural network because it operates over a volume of inputs. We configured the architecture of CNN as a multi-layer network that is designed to require minimal data processing. Sequentially five layers were designed with the following hyper-parameters. The first layer is a one-dimensional convolutional layer of 64 filters, two kernels and a Relu activation function. We add a dense (16 units), a max pooling layer and a flattened layer into the model. The output layer contains 'softmax' activation and the number of output classes.

3. Symptoms reduction

Understanding emerging symptoms is very important in disease prediction. However, symptoms are complex and could co-occur due to many diseases. Using PCA techniques, we reduce the number of symptoms to detect the degree of essential features as symptoms for each disease. The principal components are eigenvectors of the data's covariance matrix. Our data set is expressed by the matrix $X \in \mathbb{R}^{n\times d}$, and the calculation for the covariance matrix can be expressed as:

$$C = \frac{1}{n-1}\sum_{i=1}^{n}(X_i - \bar{X})(X_i - \bar{X})^T \tag{2}$$

We also determine similar diseases based on common symptoms based on a distance function to understand the ambiguity of the disease that is represented by symptoms. We use a cosine distance function as follows:

$$similarity(A,B) = \frac{A \cdot B}{\|A\| \times \|B\|} = \frac{\sum_{i=1}^{n} A_i \times B_i}{\sqrt{\sum_{i=1}^{n} A_i^2} \times \sqrt{\sum_{i=1}^{n} B_i^2}} \tag{3}$$

where A and B are vectors or matrices of diseases with symptoms.

Based on the similarity metrics, K-means clustering algorithm identifies similar diseases. The objective of K-Means clustering is to minimize total intra-cluster variance or the squared error function:

$$J = \sum_{j=1}^{K}\sum_{n\in S_j}\left|x_n - \mu_j\right|^2, \tag{4}$$

where k is a number of clusters, j is a number of instances, $x_n - \mu_j$ is an euclidean distance between $x$ instance and the j$^{th}$ cluster centroid.

The Silhouette score is used to measure the degree of separation between clusters. A score of 1 denotes the best clustering with excellent separation. The value of the silhouette coefficient is between [-1, 1].

$$Silhouette - score = \frac{b_i - a_i}{\max(bi, a_i)} \quad (5)$$

where $a_i$ - average distance between i and all of other points in its own cluster, $b_i$ - distance between i and its next nearest cluster centroid.

**RESULTS**

We first build a network of diseases and symptoms to understand predictability. The network is visualized in Figure 1. The network is modeled to have a diameter of 10, a radius of 5, and an average shortest path length of 4.2. The average disease-symptom linkage degree of the disease-symptom correlation network is 1.882.

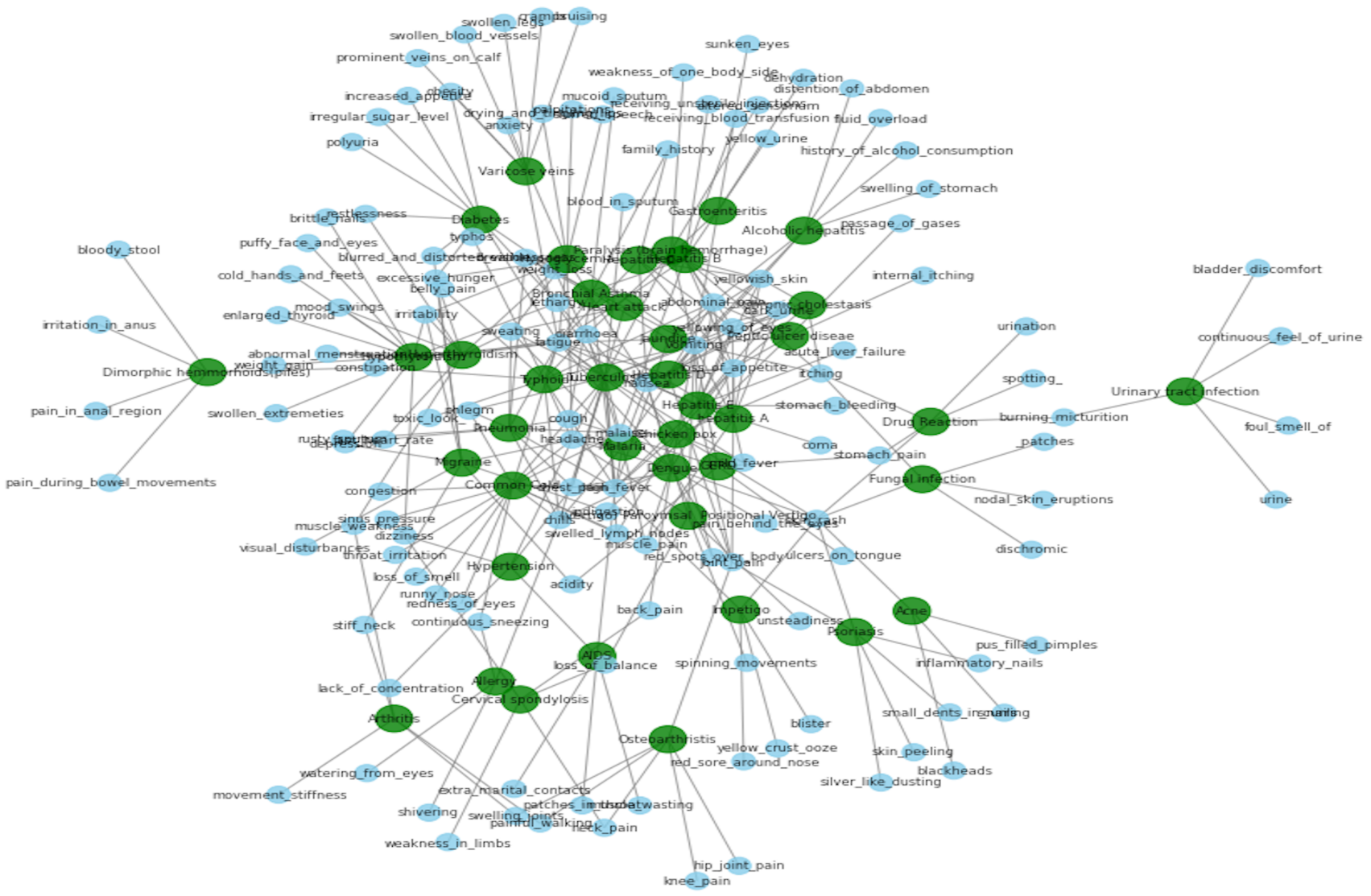

Figure 1. Network of disease and symptom relation

We identified common and unusual symptoms based on the symptoms occurring in diseases, as shown in Figure 2. Unusual symptoms were not observed in some diseases, such as Chickenpox, Chronic cholestasis, Heart attack, Jaundice, Malaria, Hepatitis A, C, D, and Hyperthyroidism.

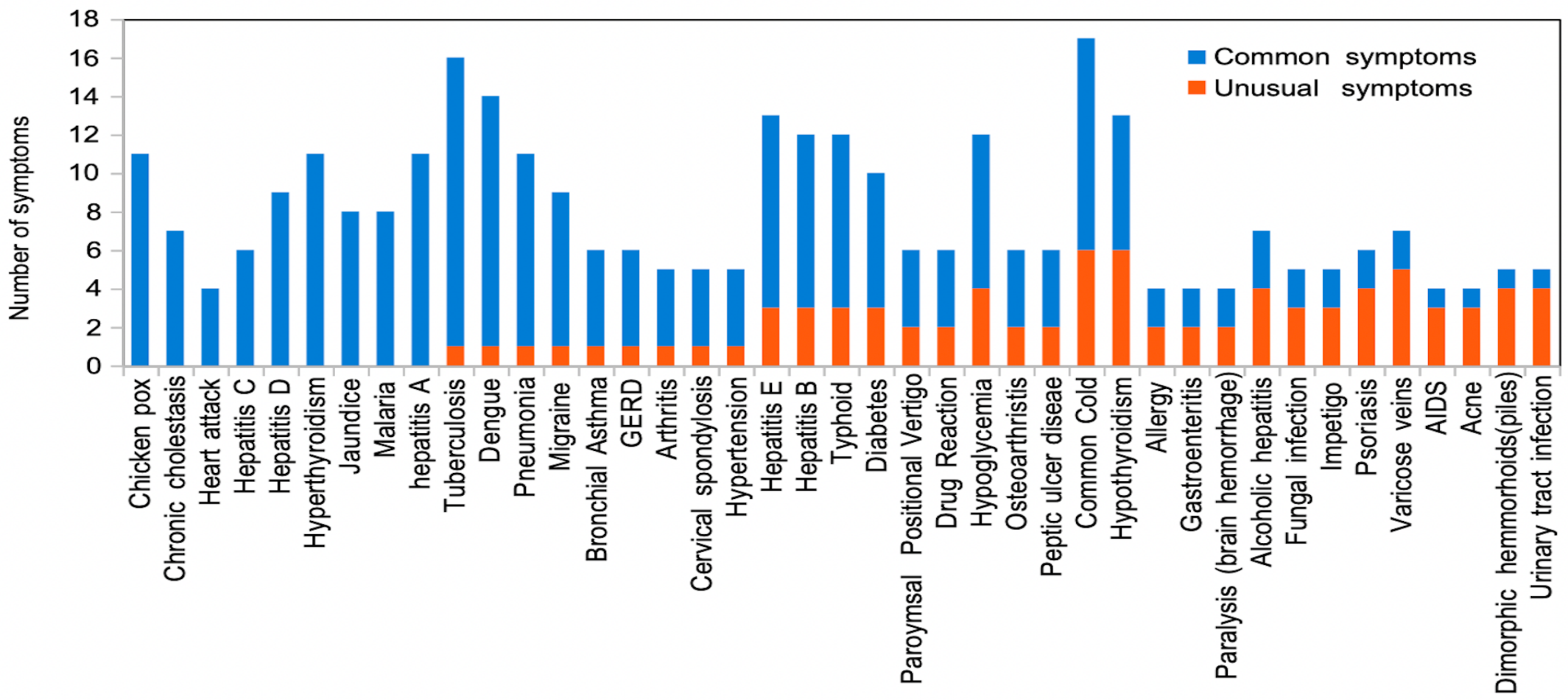

Figure 2. Common vs Unusual symptoms occurring in diseases

The occurrence of symptoms over diseases are estimated in Figure 3. A symptom occurring only in a single disease is 84, a symptom occurring in 2 diseases is 20, and so. Contrary to unusual symptoms, some symptoms occur commonly. For instance, a symptom occurring in 17 diseases is 2.

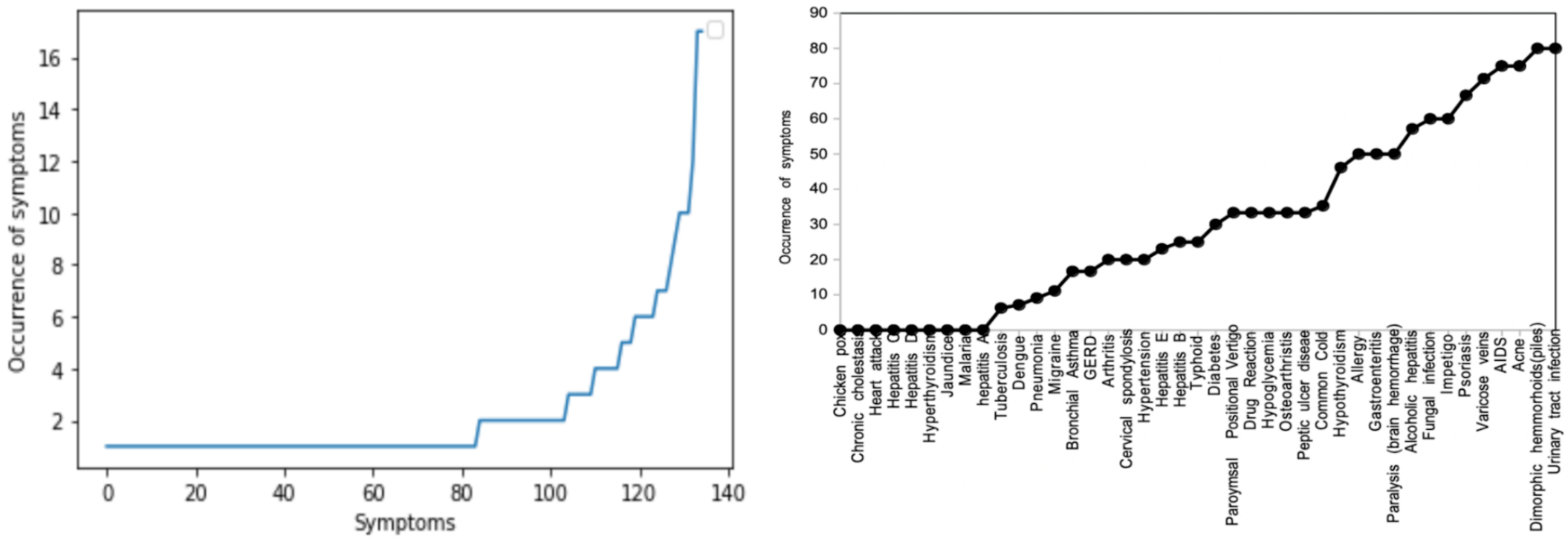

(a) (b)

Figure 3. a) Occurrence of symptoms over diseases b) Occurrence of unusual symptoms by diseases

More than two occurrences of symptoms in the disease are almost 50% of the symptoms. That highlights that most symptoms commonly occur over disease, and ambiguity of diseases is due to the common symptoms. On average, the rate of unusual symptoms occurrence for each disease is around 39.2%, except for the diseases without any unusual symptoms. The diseases with more than 50% of the uniqueness rate of symptoms are predicted relatively well.

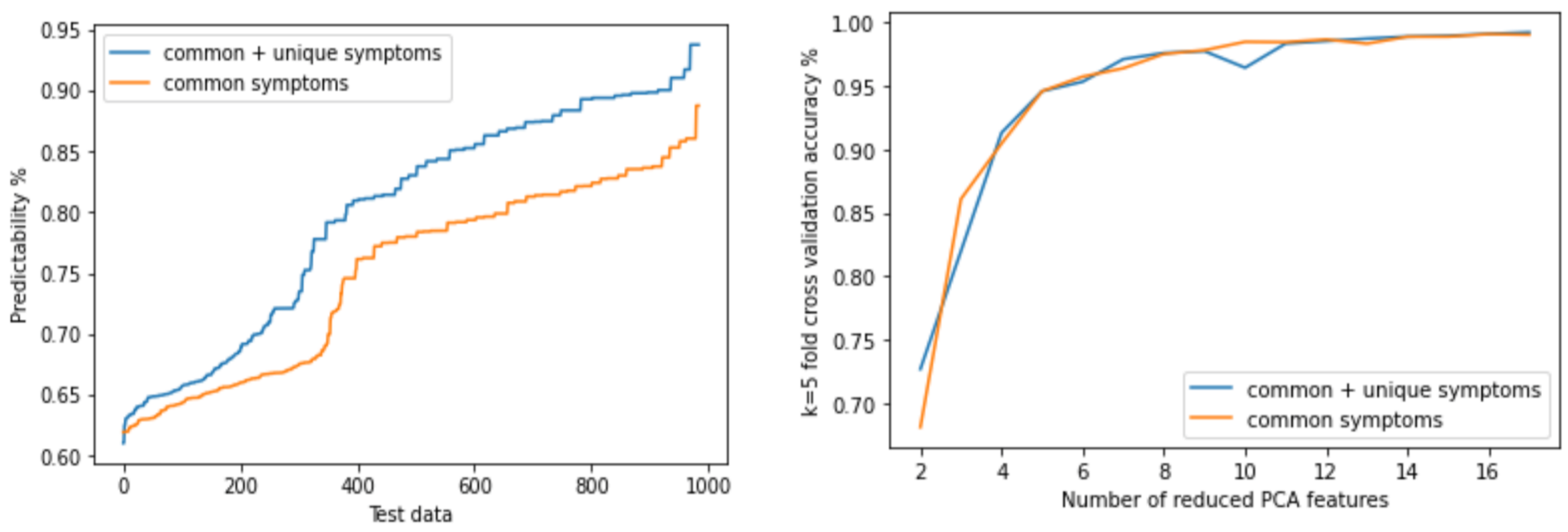

**(a)** **(b)**

Figure 4. a) Predictability rate of SVM b) K-fold cross validation using a reduced number of PCA features

We trained the disease prediction models that employ CNN and SVM methods in our split data 80/20. The dataset is well-balanced. The performance of the models are evaluated using F1-score, Precision, and Recall. The performance result in macro averaged metrics presented in Table 1 explains the predictability of diseases. Given symptoms, predicting a particular disease is 100% of the F1-score. The evaluation results were relatively good, 98% - 100% considering common and unusual symptoms. These results indicate that machine learning can potentially diagnose diseases in the early stage, taking the characteristics of symptoms into account. Our model outperformed the other experiments [8] using the same dataset, with a 3-5% increase. We also highlight that unusual symptoms increase the accuracy of the disease prediction task. Such a result was also validated by estimating the prediction probabilities of the SVM model on the test data, as shown in Figure 4.

Table 1. Evaluation result of machine learning models

| Algorithms | F1-score | Precision | Recall |
|---|---|---|---|
| SVM (common symptoms) | 98.2% | 98.4% | 98% |
| SVM (common + unusual symptoms) | 99.2% | 99.2% | 99.2% |
| CNN (common symptoms) | 99% | 99% | 99% |
| CNN (common + unusual symptoms) | 100% | 100% | 100% |

The predictability percentage increases by the prediction SVM model considering both common and unusual types of symptoms. However, unusual symptoms were important to diagnose a particular disease, but the F1-score prediction for each disease was relatively high, around 100%. However, the F1-score of the following diseases is relatively lower, as Acne disease - 81.5%, Impetigo - 87.5%, Paralysis (brain hemorrhage) - 88.8%, Paroymsal Positional Vertigo - 91.6%, Urinary tract infection - 93.7%, Hepatitis E - 96.7%, Hepatitis D - 97.7%, Gastroenteritis - 97.8%, Fungal infection - 98.1%, Psoriasis - 98.7%. Those diseases having a lower F1-score were not correlated to the uniqueness of symptoms and the number of symptoms. The reason could be related to the insufficient dataset.

However, understanding common and unusual symptoms are essential in disease prediction tasks; we try to reduce the number of symptoms employing Principal Component Analysis (PCA). Figure 4b shows the reduced number of symptoms compared to the accuracy of the SVM model. At least four types of symptoms, regardless of common or unusual characteristics, should be defined for each disease to have a predictability rate of more than 91% by a k (k=5) fold cross-validation.

That means if patients can observe at least four types of symptoms, the likelihood of disease diagnosis increases to 91%. However, diseases based on symptoms characteristic were well classified, but we need to understand the ambiguity of common symptoms to refine the classification task. So, identifying similar diseases which show similar symptoms is crucial. By K-means, diseases were clustered based on common symptoms with the use of the cosine distance method for the similarity of diseases. Comparing 41 types of diseases pairwise, 50% of diseases, around 400 pairs, have entirely different symptoms.

Very similar 20 diseases were observed and contained common symptoms among those diseases. The symptoms co-occur, and similarities of diseases increase the risk of misdiagnosing this type of disease, considering only common symptoms. Therefore, it is necessary to conduct a detailed examination using other testing equipment like blood tests, X-rays, CTs, and so. We also validated the result using the K-means clustering algorithm. The K-means method encourages us to identify similar diseases based on common symptoms. We applied K-means clustering with a cosine distance. Figure 5 presents the Silhouette score of K-means on the data containing PCA-reduced symptoms and all symptoms. PCA reduction improved K-means performance as the Silhouette score is higher than 0.5 on the dataset containing PCA-reduced symptoms. The Silhouette score is 0.64 when the optimal cluster number is 6 for the dataset containing PCA-reduced symptoms; the Silhouette score is 0.28 when the optimal cluster number is 7 for the dataset containing all symptoms. From cluster 27, the silhouette score curve became flattened at a score of 0.74.

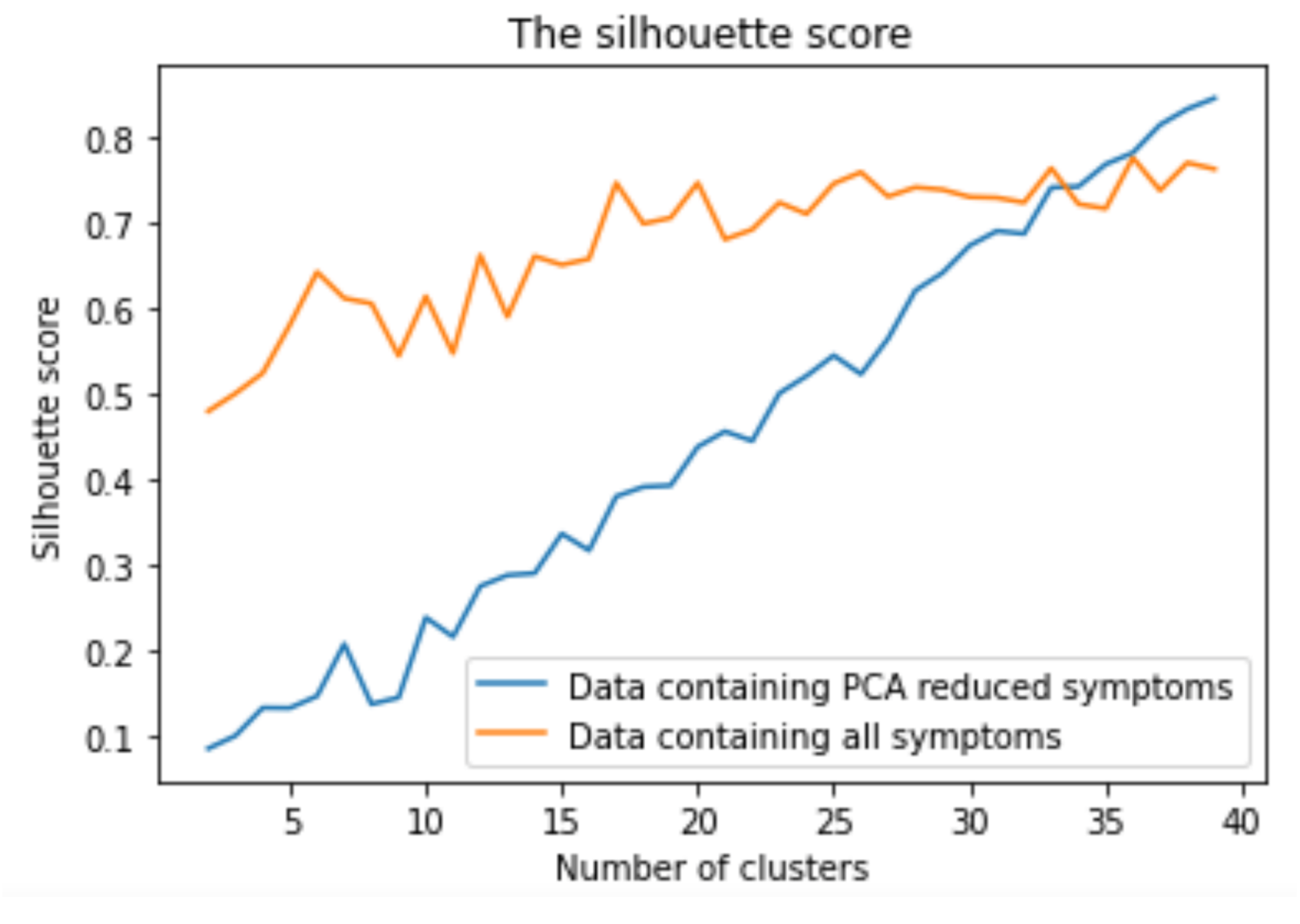

Figure 5. Silhouette score of K-means clustering

The result indicates that the dataset containing PCA-reduced symptoms was better classified, and similar diseases were identified. However, the dataset containing all symptoms got a silhouette score of 0.84 at cluster 40. The result explains that most diseases were well separated given the symptoms and why the machine learning method performed well. Table 2 summarizes the result of the K-means clustering.

Table 2. K-means clustering on disease-symptoms dataset

| | Dataset containing all symptoms | Dataset containing PCA reduced symptoms |
|---|---|---|
| Clusters | k1=840 instances<br>k2=840 instances<br>k3=360 instances<br>k4=2160 instances<br>k5=120 instances<br>k6=480 instances<br>k7=120 instances | k1=600 instances<br>k2=594 instances<br>k3=480 instances<br>k4=840 instances<br>k5=2046 instances<br>k6=360 instances |
| Silhouette score | 0.28 | 0.64 |

## CONCLUSIONS

This study analyzed 41 types of diseases, including 135 common and uncommon symptoms of patients. We can use machine learning methods to diagnose a particular disease given symptoms with 98-100% accuracy. Our results indicate that unusual and uncommon symptoms increase disease prediction accuracy. However, most diseases have common symptoms and could co-occur; it is difficult to understand the crucial features of symptoms. Our results suggest that data reduction techniques allow import features of symptoms by reducing the number of symptoms. In our demonstration, at least four types of symptoms for each disease were sufficient to diagnose a particular disease with more than 91% of accuracy. To analyze the ambiguity of diseases, we need to identify similar diseases based on common symptoms. However, a large dataset is required for analyzing the relationship between diseases and symptoms in a wide range, such as the asynchronous onset of diseases. We will extend this study by using open databases of biomedical protein, molecular, gene, and phenotypic databases and extracting information from clinical article databases (PubMed). The relationship between the clinical manifestations of the disease and their underlying molecular interactions based on symptoms will be explored in detail.

**Acknowledgements:** This research was supported by Mongolian Foundation for Science and Technology (MFST), project number "STIPICD-2021/475". The study was supported in part by a grant from Irkutsk National Research Technical University. The authors would like to give special thanks to the Mongolian Ministry of Science and Technology, and the Irkutsk National Research Technical University for supporting this research.

**Competing interests** The authors declare no conflict of interest. The funders had no role in the design of the study; in the collection, analyses, or interpretation of data; in the writing of the manuscript; or in the decision to publish the results.

**Patient consent for publication** Not applicable.

**Data availability statement** Data may be obtained from a third party and are not publicly available.

**Conflicts of Interest:** The authors declare no conflict of interest. The funders had no role in the design of the study; in the collection, analyses, or interpretation of data; in the writing of the manuscript; or in the decision to publish the results.